\documentclass{article}

\usepackage{microtype}
\usepackage{graphicx}
\usepackage{subcaption}
\usepackage{booktabs} 

\usepackage{hyperref}

\usepackage[accepted]{icml2026}

\usepackage{amsmath}
\usepackage{amssymb}
\usepackage{mathtools}
\usepackage{amsthm}

\usepackage[capitalize,noabbrev]{cleveref}

\theoremstyle{plain}

\theoremstyle{definition}

\theoremstyle{remark}

\usepackage[disable,textsize=tiny]{todonotes}

\icmltitlerunning{Offline Supervision for Reinforcement Learning in VLA Models}

\begin{document}

\twocolumn[
  \icmltitle{Leveraging Offline Supervision for Efficient and Generalizable Reinforcement Learning in Large-Scale Vision--Language--Action Models}



  \icmlsetsymbol{equal}{*}

  \begin{icmlauthorlist}
    \icmlauthor{Dmitriy Poyarkov}{comp}
    \icmlauthor{Aleksei Staroverov}{yyy,comp}
    \icmlauthor{Aleksandr I. Panov}{yyy,comp}
  \end{icmlauthorlist}

  \icmlaffiliation{comp}{AXXX}
  \icmlaffiliation{yyy}{MIRAI}

  \icmlcorrespondingauthor{Dmitriy Poyarkov}{chmod\_700@outlook.com}
  \icmlcorrespondingauthor{Aleksei Staroverov}{alstarmmm@gmail.com}

  \icmlkeywords{Machine Learning, ICML}

  \vskip 0.3in
]



\printAffiliationsAndNotice{}  

\begin{abstract}
  It is commonly observed that online reinforcement learning (RL) produces better-performing strategies than offline methods across a broad range of performance measures. In particular, RL-trained policies exhibit stronger out-of-distribution (OOD) behavior, where models trained only with imitation learning approaches often struggle. A recent study introduced an OOD-focused benchmark and reported that RL-trained vision--language--action (VLA) policies achieve noticeably better OOD performance and slightly better in-distribution (IND) performance than their counterparts trained with supervised fine-tuning (SFT). In this work, we investigate whether hybrid offline--online training can combine the advantages of both approaches. Specifically, we study RL methods regularized by offline supervision via either offline data or an offline-trained reference policy. We evaluate these approaches on the OOD benchmark and compare them with both offline-only training and standard RL. Our results show that although offline training achieves limited OOD performance by itself, incorporating offline supervision into RL preserves strong OOD capability while substantially improving training efficiency. In particular, the guided methods reach performance close to that of standard RL while requiring roughly half of the training budget. Rather than producing a trade-off between speed and OOD performance, the hybrid approach retains strong OOD capability while achieving this efficiency gain. Project page: \url{https://alstar8.github.io/offline-supervision-vla-rl/}.
\end{abstract}

\section{Introduction}

Machine learning (ML) policies can be trained either offline, using fixed
datasets, or online, through interaction with the environment. In
practice, online reinforcement learning (RL) often produces stronger
final policies, especially when robustness and adaptation are important.
At the same time, offline training remains substantially cheaper and
simpler to deploy, and is therefore the default option in many realistic
robotic settings. This gap is also reflected in prior work on imitation
learning and learning from demonstration, where purely offline approaches
often struggle more under distribution shift
\cite{argall2009lfd,ross2011dagger,ren2021generalization}.

\begin{figure}[t]
\centering
\includegraphics[width=\linewidth]{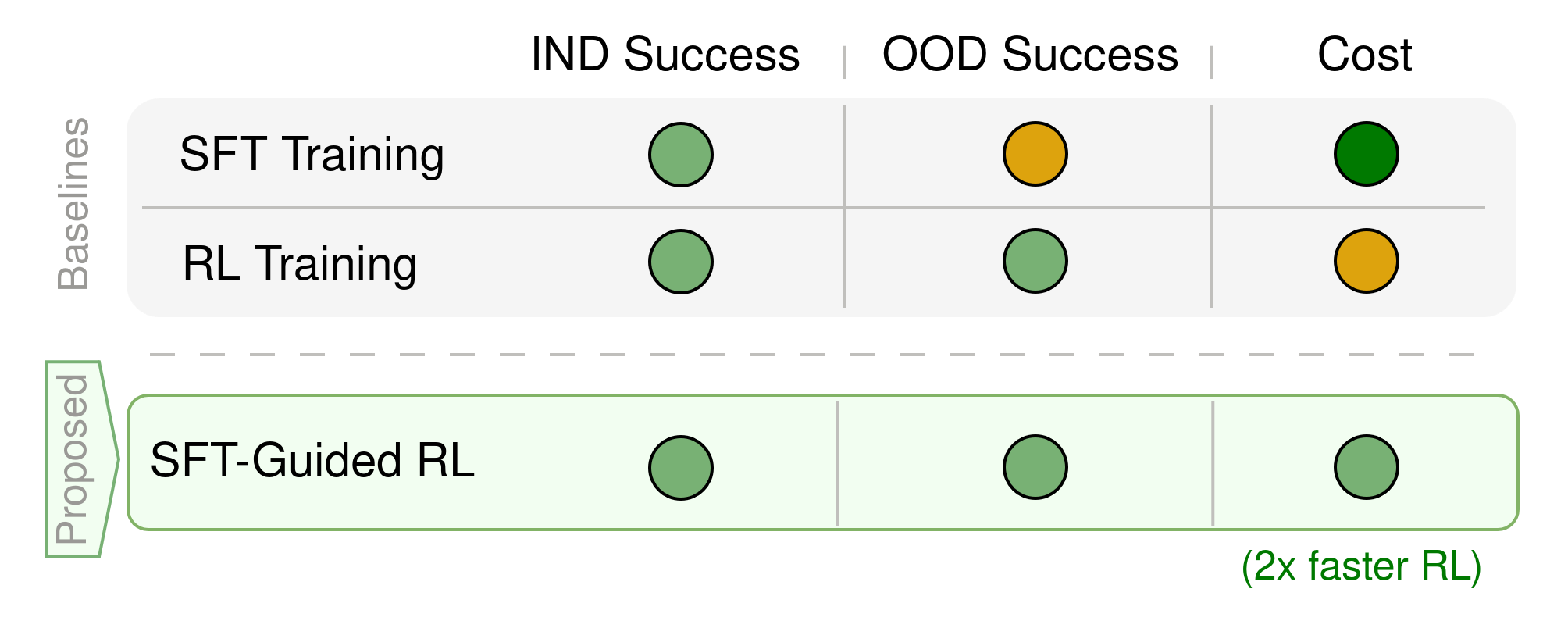}
\caption{The core idea of combining offline and online training in the described setting.}
\label{fig:idea}
\end{figure}

This trade-off is particularly visible in large-scale
vision--language--action (VLA) models, including RT-1, RT-2, Octo,
$\pi_0$, $\pi_{0.5}$, OpenVLA, and related systems
\cite{brohan2022rt1,zitkovich2023rt2,octo2024,black2025pi0,physicalintelligence2025pi05,kim2024openvla}. Recent work has shown that RL fine-tuning of OpenVLA low-rank adaptation (LoRA) adapters \cite{kim2024openvla,hu2021lora} can improve out-of-distribution (OOD) behavior relative to supervised fine-tuning (SFT), but with a much higher online training cost \cite{liu2025what}. This motivates a more specific question: in parameter-efficient VLA adaptation, can offline supervision be used not only for initialization, but as a structured training signal that makes RL optimization more efficient?

Hybrid offline--online methods are an active area of research
\cite{nair2020accelerating,nakamoto2023calql,lee2022offline2online,kang2018pofd}, but their
behavior is not uniform across model classes and training regimes. In
particular, low-rank adaptation of large transformer-based VLAs is not a
standard pretrain--finetune setting: the base model remains frozen, only
a small adapter is optimized, and the interaction between supervised and
RL objectives may differ substantially from that of conventional policy
networks \cite{hu2021lora,hancock2025actions}. As a result, whether common hybrid strategies transfer to this
setting, and which forms of offline guidance are most effective, remains
an open question. The overall motivation and training setup studied in
this paper are illustrated in Fig.~\ref{fig:idea}.

In this work, we study whether offline supervision can be incorporated
into RL fine-tuning of large-scale VLA policies in a way that improves
training efficiency without sacrificing the robustness obtained from
online adaptation. We consider two guided variants of Proximal Policy Optimization
(PPO): one that regularizes the policy toward a frozen SFT reference
model through a Kullback--Leibler (KL) penalty, and
one that augments PPO with behavior cloning on the offline
demonstration dataset. As a simple baseline, we also evaluate PPO
initialized directly from the SFT adapter. The focus of the paper is
therefore not only whether guided training performs better, but also
what this reveals about the role of supervised priors during RL
optimization of LoRA-based VLA policies.

Our central finding is that offline guidance serves as an effective
optimization prior in this setting. Both guided variants improve over
standard PPO at the same online training budget, and the reference-guided
variant reaches nearly the same operating point as a substantially longer
PPO run while using far fewer environment interactions.

Our contributions are as follows:
\begin{itemize}
\item We study hybrid offline--online training for LoRA-based RL
fine-tuning of large-scale VLA models and formulate two simple
offline-guided PPO variants based on reference-policy regularization and
offline behavior cloning.
\item We show, in a controlled OpenVLA fine-tuning setting evaluated with
both in-distribution and out-of-distribution tests, that offline guidance
can improve training efficiency while preserving the benefits of RL-based
adaptation.
\item We provide a broader analysis of the proposed hybrid training
setting, including comparisons between guidance variants, sensitivity to
regularization design choices, and additional observations about the
resulting optimization behavior.
\end{itemize}

\begin{figure}
    \centering
    \includegraphics[width=1\linewidth]{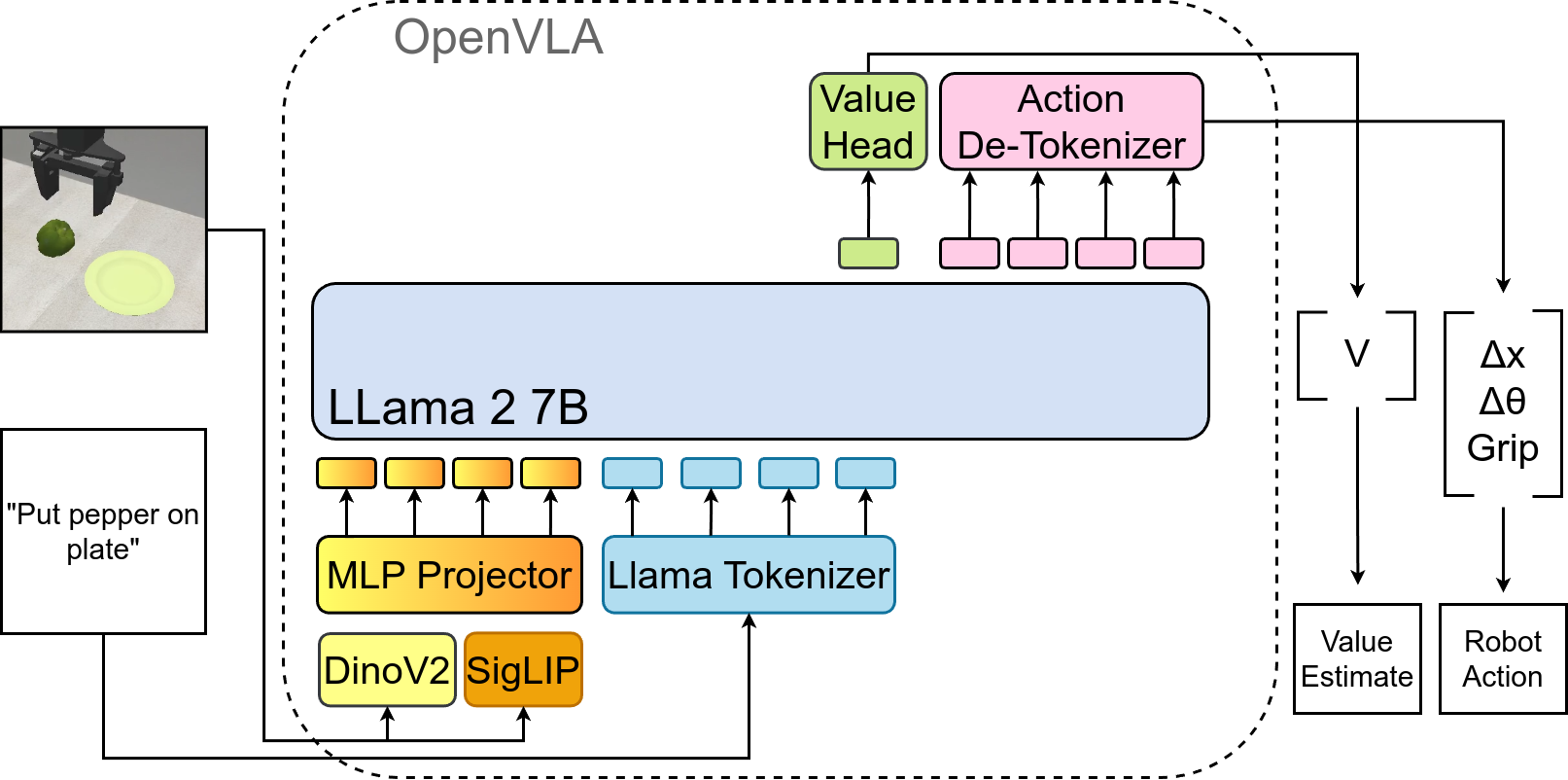}
    \caption{Architecture of the OpenVLA model used in our setting. The architecture is consistent with the base OpenVLA model and the value-head modification.}
    \label{fig:openvla_architecture}
\end{figure}

\section{Related Work}

\subsection{Reinforcement Learning for Vision--Language--Action Models}

Recent work has begun to study RL as a means of improving the generalization of VLA policies beyond what is obtained from SFT alone. In particular, \cite{liu2025what} provides evidence that PPO-based RL fine-tuning \cite{schulman2017ppo} can yield substantially stronger semantic and execution OOD behavior than SFT, while requiring significantly more expensive online training. This establishes a central trade-off between training efficiency and robustness in VLA models and motivates the hybrid training question studied here.

Our setting uses OpenVLA \cite{kim2024openvla}, an open-source large-scale VLA model that predicts tokenized robot actions from image and language inputs. More broadly, it belongs to a recent class of generalist VLA policies that includes RT-1, RT-2, Octo, $\pi_0$, and $\pi_{0.5}$ \cite{brohan2022rt1,zitkovich2023rt2,octo2024,black2025pi0,physicalintelligence2025pi05}. A schematic overview of the architecture used in our setting is shown in Fig.~\ref{fig:openvla_architecture}.

\subsection{Hybrid RL Training with Teacher Guidance}

Several works have explored combining reinforcement learning with guidance from a reference or teacher policy. \cite{schmitt2018kickstarting} proposes a teacher--student framework in which a student policy is trained with RL while being regularized toward a stronger teacher policy through a Kullback--Leibler (KL) divergence term. This mechanism allows the student to benefit from the teacher's prior knowledge while still improving through interaction.

A closely related formulation appears in Proximal Policy Distillation (PPD) \cite{spigler2024proximal}. PPD augments the PPO objective with a KL divergence to a teacher policy evaluated on the same minibatches used for RL updates. This regularization stabilizes optimization and enables knowledge transfer from a pretrained policy while maintaining the benefits of on-policy RL optimization.

Regularization toward a pretrained reference policy is also central to reinforcement learning from human feedback (RLHF). In \cite{ouyang2022training}, the policy is first initialized via supervised fine-tuning and subsequently optimized with PPO against a learned reward model. During PPO training, the objective includes a KL penalty to the supervised reference model, which constrains policy updates and limits drift from the pretrained behavior. Some variants additionally mix in supervised gradients from the pretraining distribution during RL optimization to further preserve pretrained capabilities.

\subsection{Catastrophic Forgetting in Offline-to-Online Fine-Tuning}

A major difficulty in combined offline--online training is catastrophic forgetting: during RL optimization the policy may lose behaviors acquired during pretraining or supervised fine-tuning before RL has recovered them \cite{kirkpatrick2017ewc}. This phenomenon has been identified as a central failure mode in RL fine-tuning, particularly when the online objective provides a narrow training signal relative to the diversity of behaviors captured by the pretrained model.

\cite{baker2022vpt} address this issue in Video PreTraining (VPT) by regularizing the RL policy toward the pretrained model using a KL constraint. Their ablations show that removing this regularization leads to training collapse after initial progress, which they attribute to catastrophic forgetting of previously learned behaviors.

In the context of VLA models, \cite{hancock2025actions} show that parameter-efficient adaptation can substantially mitigate catastrophic forgetting. Their work demonstrates that LoRA-based fine-tuning can preserve the underlying vision-language model capabilities while adapting it into a VLA policy, avoiding the large capability drop often observed with full fine-tuning. This observation is further supported by \cite{liu2025what} and is consistent with our own experimental findings.

\subsection{Summary}

Prior work highlights a tension between training efficiency and generalization in policy learning. RL-based training can improve robustness relative to supervised fine-tuning alone \cite{liu2025what}, but requires substantially more expensive online training. Several works therefore explore hybrid objectives that combine RL updates with guidance from pretrained or reference policies in order to stabilize training and leverage prior knowledge \cite{schmitt2018kickstarting,spigler2024proximal,ouyang2022training}. Our work investigates this direction in the context of parameter-efficient VLA RL training.

\section{Task Definition}

We model the problem as a partially observable Markov decision process
(POMDP). At each step $t$, the agent receives an observation
$o_t = (I_t, l)$ consisting of the current RGB image $I_t$ and a natural
language instruction $l$. The instruction remains fixed within an
episode, but is provided to the policy at every step together with the
current image. Let $x_t \in \mathcal{X}$ denote the underlying
environment state, which is not directly observed.

The policy predicts tokenized actions $u_t \in \mathcal{U}$, which are
converted by the environment wrapper into executed robot commands
$a_t \in \mathcal{A}$. Episodes have a fixed horizon $T=80$, so PPO
rollouts are truncated by timeout rather than by a task-specific
termination condition. A trajectory has the form
$\tau = (o_0,u_0,a_0,r_0,\dots,o_T)$.

The reward is a sparse difference-based signal derived from task
progress. The agent is rewarded for increases in task progress,
rather than for repeatedly remaining in an already achieved state.

\section{Method}

This section describes the training objectives used in our experiments.
All methods operate on the same pretrained base policy architecture.
During training, we keep the base model frozen and optimize a
LoRA module \cite{hu2021lora} attached to the policy network.
The different methods considered in this work vary only in their
training objectives and the data used for optimization.

\subsection{Supervised Fine-Tuning}

We first train a base policy using supervised fine-tuning on an offline
demonstration dataset $\mathcal{D}_{\text{off}}=\{(o_i,a_i^\star)\}$.
The policy $\pi_\theta(a|o)$ is trained to maximize the likelihood of expert
actions:

\begin{align}
\mathcal{L}_{\text{SFT}}(\theta)
=
-\mathbb{E}_{(o,a^\star)\sim\mathcal{D}_{\text{off}}}
\log \pi_\theta(a^\star|o).
\end{align}

This is a standard training procedure, and we replicate it in our study to obtain a LoRA that will be used as a frozen reference policy during RL optimization.

\subsection{Proximal Policy Optimization}

We use PPO \cite{schulman2017ppo} as the baseline algorithm. The
objective used in our experiments is
\begin{align}
\mathcal{L}_{\mathrm{PPO}}(\theta)
&=
\mathbb{E}_{t}
\left[
\min
\Big(
r_t(\theta)\hat{A}_t,
\text{clip}(r_t(\theta),1-\epsilon,1+\epsilon)\hat{A}_t
\Big)
\right],
\\
r_t(\theta)
&=
\frac{\pi_\theta(a_t|o_t)}
{\pi_{\theta_{\mathrm{old}}}(a_t|o_t)} .
\end{align}

Here $\hat{A}_t$ denotes the advantage estimate and $\epsilon$ is the PPO
clipping parameter.

\subsection{Supervised Fine-Tuning Guided PPO}

\begin{figure}
    \centering
    \includegraphics[width=\linewidth]{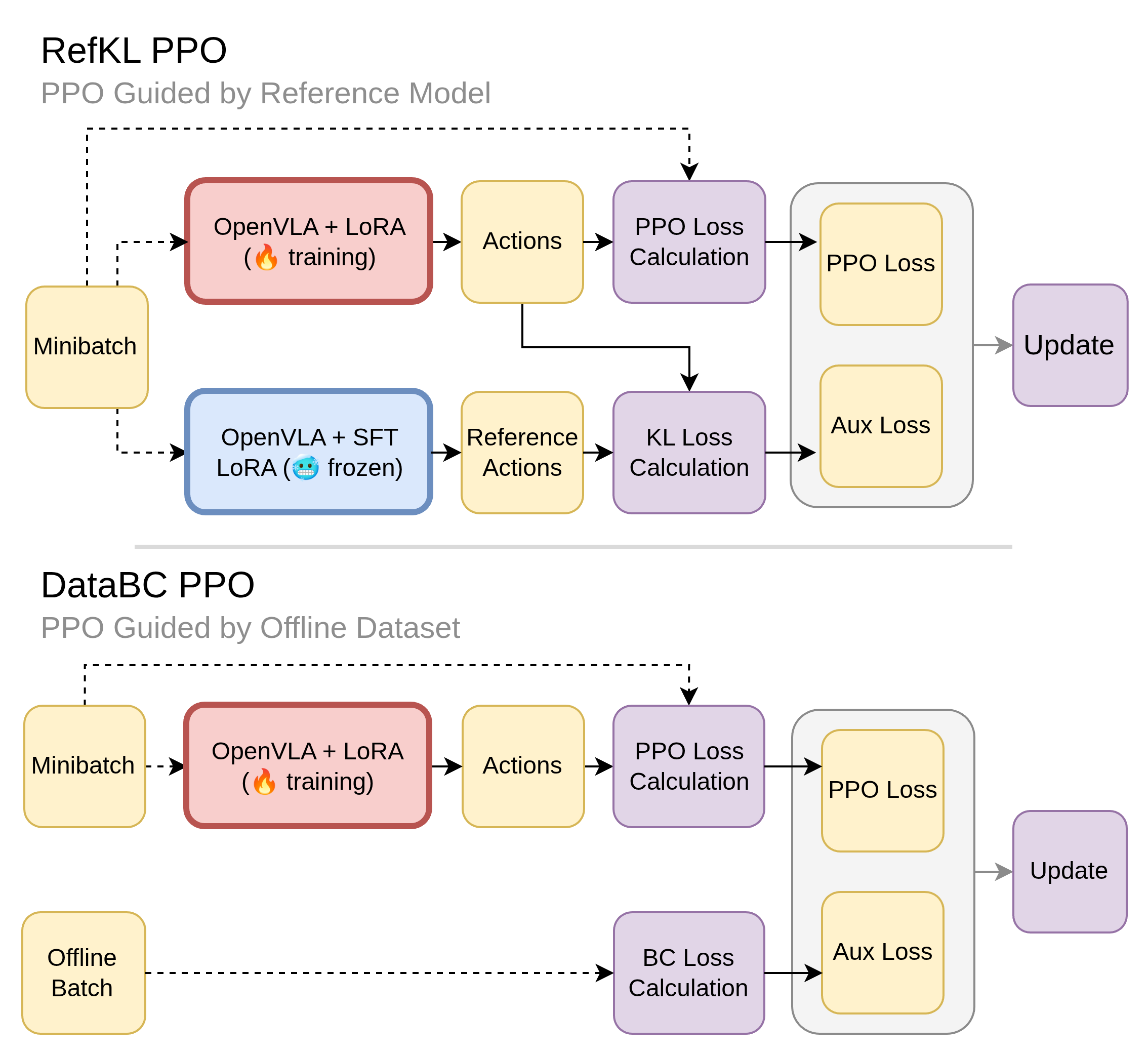}
    \caption{Training pipelines for PPO modifications used in our experiments}
    \label{fig:algs_scheme}
\end{figure}

For our SFT-guided RL, we supplement the PPO objective with an auxiliary loss term. We evaluate two main ways of doing this: via the frozen reference model (PPO with a KL penalty to the reference model, denoted RefKL) and with behavior cloning (BC) over the offline batch from the SFT dataset. The resulting training pipelines are shown in Fig.~\ref{fig:algs_scheme}. In our setting, the reference policy $\pi_{\mathrm{ref}}$ is the frozen
SFT model. In other words, we use either the offline data or the corresponding model trained on that data.

\paragraph{PPO Guided by a Reference Model}

The policy is regularized to imitate the reference policy by
penalizing the KL divergence between the current policy and the
reference policy on the sampled minibatch. We refer to this variant as
RefKL:
\begin{align}
\mathcal{L}_{\mathrm{RefKL}}(\theta)
=
\frac{1}{|\mathcal{B}|}
\sum_{o \in \mathcal{B}}
\mathrm{KL}\big(
\pi_{\mathrm{ref}}(\cdot|o)
\;\|\;
\pi_{\theta}(\cdot|o)
\big).
\end{align}

The full optimization objective becomes
\begin{align}
\mathcal{L}(\theta)
=
\mathcal{L}_{\mathrm{PPO}}(\theta)
+
\beta \mathcal{L}_{\mathrm{RefKL}}(\theta),
\end{align}

where $\beta$ controls the strength of the regularization toward the
reference policy.

\subsection{PPO Guided by Offline Dataset (DataBC)}

In addition to reference-model regularization, we also consider
augmenting PPO with behavior cloning on the offline dataset; we denote
this variant DataBC.

Given offline demonstrations $(o,a^\star)\sim\mathcal{D}_{\text{off}}$,
we define
\begin{align}
\mathcal{L}_{\mathrm{DataBC}}(\theta)
=
-\mathbb{E}_{(o,a^\star)\sim\mathcal{D}_{\text{off}}}
\log \pi_\theta(a^\star|o).
\end{align}

The joint objective is
\begin{align}
\mathcal{L}(\theta)
=
\mathcal{L}_{\mathrm{PPO}}(\theta)
+
\beta \mathcal{L}_{\mathrm{DataBC}}(\theta).
\end{align}

\subsection{Regularization Weight Scheduling}
\label{sec:scheduling}

The auxiliary losses described above are weighted by a coefficient $\beta$.
Rather than using a fixed value throughout training, we employ a simple
curriculum schedule that gradually removes the auxiliary constraint as
training progresses.
Formally, the coefficient is defined as

\begin{align}
\beta(t) =
\begin{cases}
\beta_0, & t < t_1 \\
\beta_0 \left(1 - \frac{t - t_1}{t_2 - t_1}\right), & t_1 \le t < t_2 \\
0, & t \ge t_2
\end{cases}
\end{align}

where $t$ denotes the training step and $t_1, t_2$ mark the start and end
of the decay phase. In our experiments, we use $t_1=100{,}000$ and
$t_2=300{,}000$, so the auxiliary term is held constant during the first
100k steps, linearly annealed over the next 200k steps, and removed
entirely thereafter. The same schedule is used for both RefKL and
DataBC.

\section{Experiments}

\subsection{Baseline Results and Reproduction}

RL4VLA \cite{liu2025what} reports that reinforcement learning substantially improves policy performance compared with supervised fine-tuning, particularly in out-of-distribution evaluation. In their results, RL-trained policies achieve an average OOD success rate of approximately $0.75$, compared with roughly $0.55$ for SFT policies, while IND success increases from around $0.78$ to $0.94$.

To support the experiments in this work, we reproduced the RL4VLA training procedure \cite{liu2025what}. The original release provides the final merged checkpoint of the RL-trained model but does not include intermediate checkpoints or the corresponding LoRA adapters. Since our experiments require policies from specific training stages, we re-ran the RL training to obtain LoRA adapters for the checkpoints used in our evaluation (including the 1M and 2M step models). We also reproduced the SFT training in order to obtain the checkpoints required for our experiments.

The reproduced RL models achieve performance close to the results reported in \cite{liu2025what}. In some cases, the reproduced checkpoints are slightly higher (1--2\%) due to training stochasticity, but the overall trends remain consistent. Our reproduced SFT checkpoint differs more noticeably: the early-stopped 2k-demo model used in this paper is stronger than the published SFT number while still remaining clearly below RL in both IND and OOD evaluation. Throughout this paper, we therefore report baselines from our reproduced runs rather than directly using previously published metrics. The resulting baseline performance is summarized in Table~\ref{tab:final_checkpoints}.

\subsection{Experimental Setup}

\begin{figure}
    \centering
    \includegraphics[width=1\linewidth]{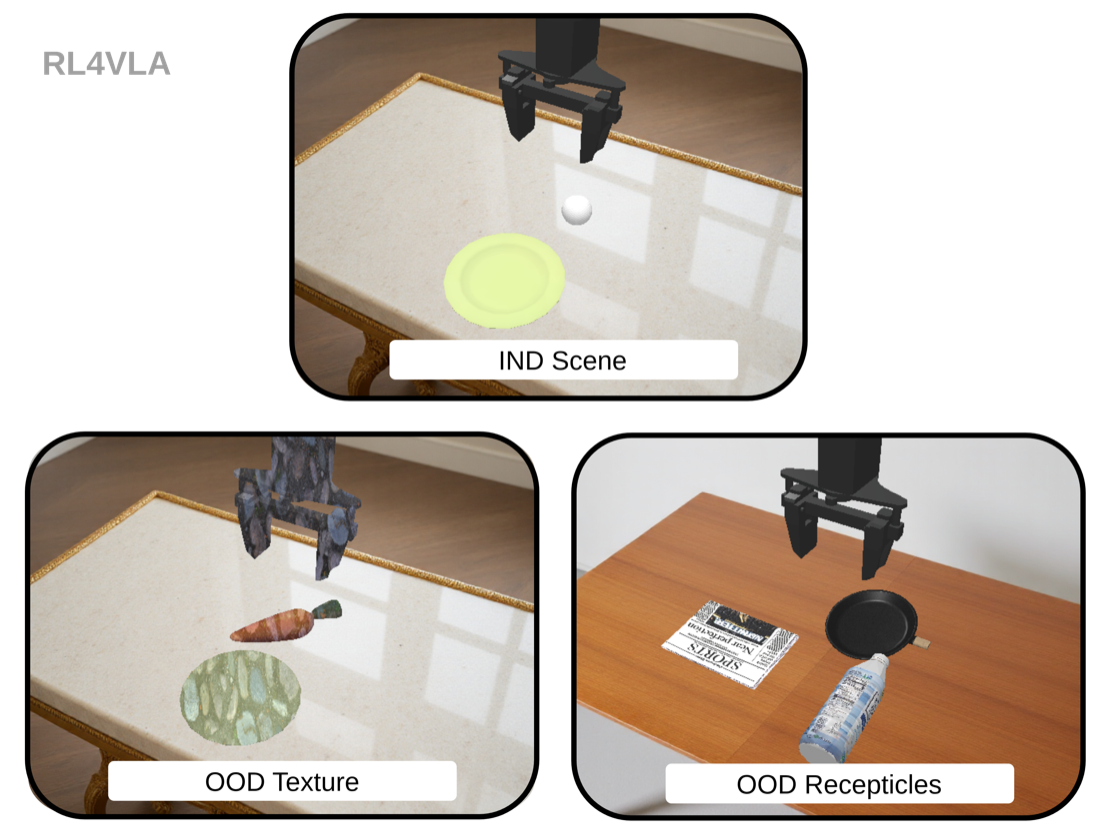}
    \caption{Examples of RL4VLA scenes. Top: an IND scene used during training. Bottom: examples of OOD scenes.}
    \label{fig:rl4vla_images}
\end{figure}

All experiments are conducted using the RL4VLA setup \cite{liu2025what}, which provides modified Simpler tasks, environments, training code, and reinforcement learning hyperparameters. Unless stated otherwise, we follow this setup and keep the original RL hyperparameters and training scale unchanged so that differences can be attributed to the training objective rather than to changes in the experimental protocol.

The benchmark provides a warm-start OpenVLA model (OpenVLA-warmup) \cite{kim2024openvla} obtained from pretraining on a small set of expert demonstration trajectories generated using the motion-planning pipeline. In all experiments, we perform RL by training a zero-initialized LoRA adapter on top of OpenVLA-warmup, while keeping the base model frozen.
For reinforcement learning, we additionally use the value-head
modification implemented in RL4VLA \cite{liu2025what}, so supervised
fine-tuning and RL share the same backbone and action interface, with
RL differing mainly through the added value prediction pathway and PPO
optimization losses.

We use the difference-based sparse reward formulation defined in RL4VLA \cite{liu2025what}. For SFT, the expert demonstration dataset is filtered to remove transitions with very small actions.

The reference implementation in \cite{liu2025what} trains the SFT policy using 16k expert demonstrations for 60k steps. In our experiments, we instead use a smaller early-stopped SFT checkpoint trained on 2k expert demonstrations for 7.5k steps. Despite the substantially smaller training budget, this checkpoint achieves stronger performance than the published SFT result and serves as the reference model for offline-guided training.

\subsection{Evaluation Protocol}

For all methods, we follow a common evaluation procedure consistent with the RL4VLA benchmark \cite{liu2025what}. Each checkpoint is evaluated using three evaluation seeds $\{0,1,2\}$. We use 64 in-distribution episodes and 960 out-of-distribution episodes in total (64 episodes for each of the 15 OOD environments; see Fig.~\ref{fig:rl4vla_images}). The OOD environments are grouped into vision, language, and action shifts. Reported success rates are averaged across both evaluation seeds and training seeds.

Unless stated otherwise, each experiment is run with two training seeds. Due to computational limitations, some configurations are reported from a single run.

\subsection{Offline-Guided PPO}

\begin{figure}[t]
    \centering
    \begin{subfigure}{0.49\linewidth}
        \centering
        \includegraphics[width=\linewidth]{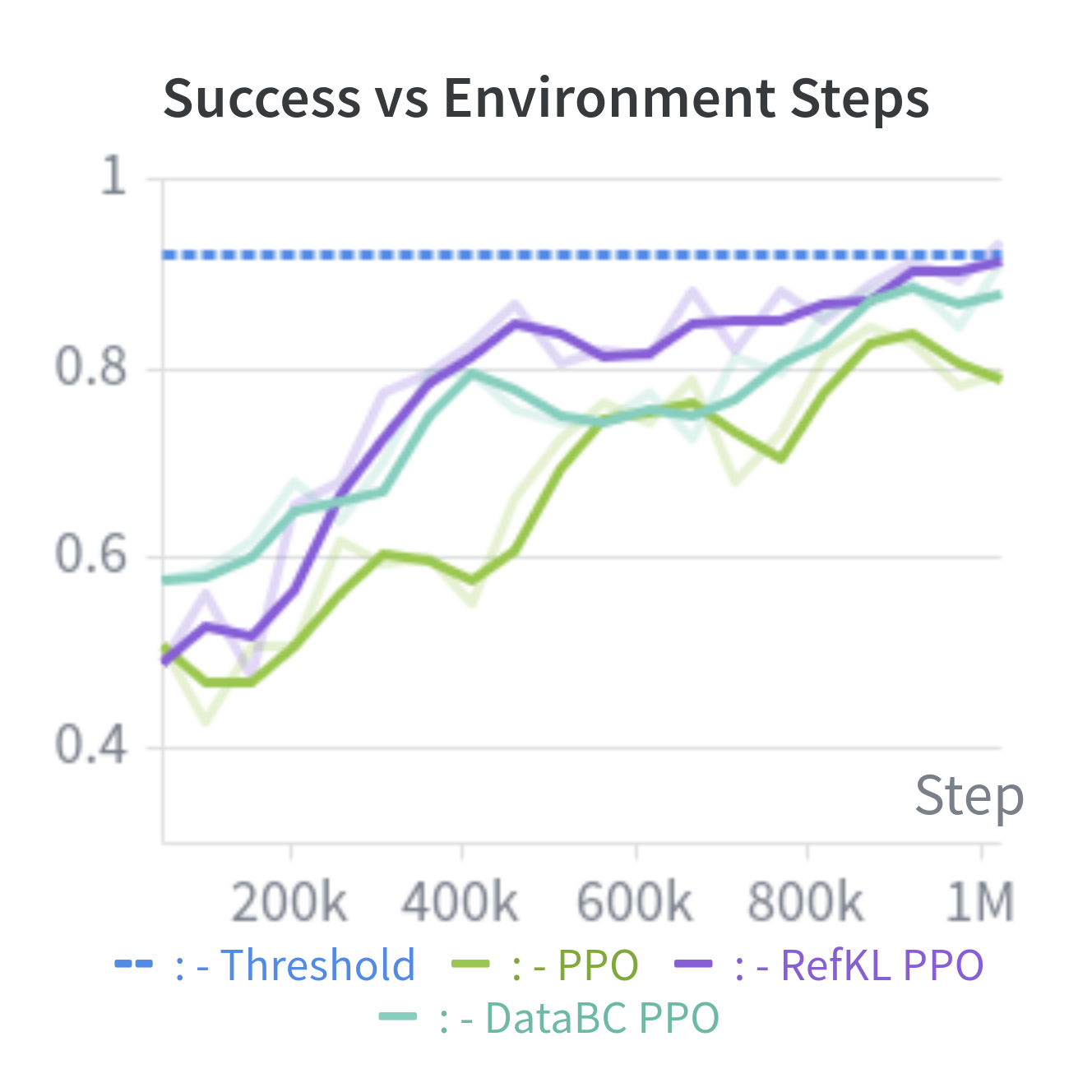}
    \end{subfigure}
    \hfill
    \begin{subfigure}{0.49\linewidth}
        \centering
        \includegraphics[width=\linewidth]{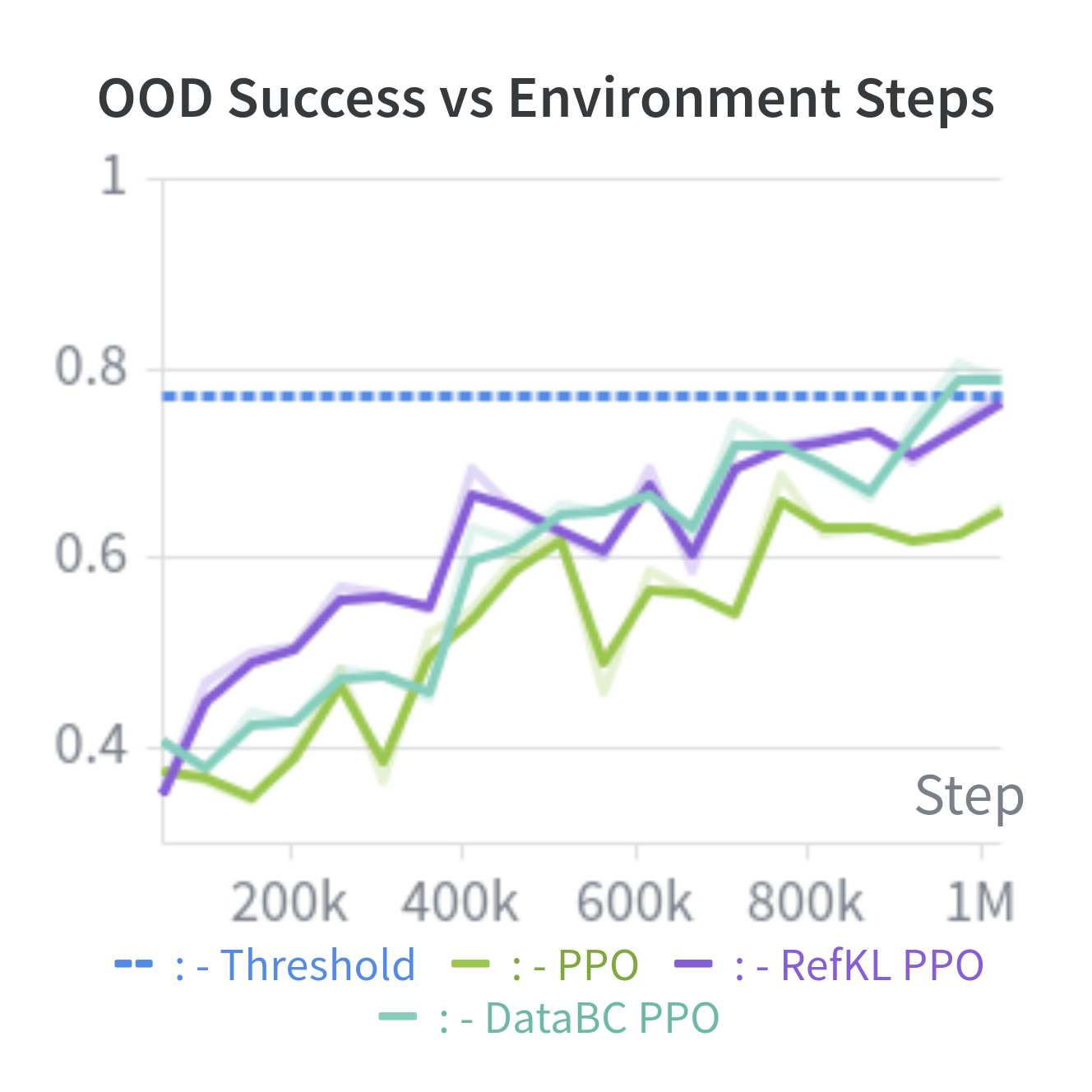}
    \end{subfigure}
    \caption{Training curves for in-distribution and out-of-distribution success. Dashed lines indicate the final PPO performance at 2M steps.}
    \label{fig:training_curves}
\end{figure}

\begin{table}[t]
\centering
\small
\setlength{\tabcolsep}{5pt}
\renewcommand{\arraystretch}{1.1}

\begin{tabular}{lccccc}
\toprule
 & IND & \multicolumn{4}{c}{OOD} \\
\cmidrule(lr){3-6}
Method &  & Act & Lang & Vis & Avg \\
\midrule
SFT & 0.82 & 0.46 & 0.60 & 0.74 & 0.62 \\

PPO SFT-init (1M)
& 0.87
& 0.51
& 0.69
& 0.78
& 0.69 \\

PPO (2M)
& 0.92
& 0.82
& 0.75
& 0.76
& 0.77 \\

PPO RefKL (1M)
& 0.93
& 0.79
& 0.76
& 0.76
& 0.77 \\

PPO DataBC (1M)
& 0.91 
& 0.74 
& 0.73 
& 0.74 
& 0.74 \\

\bottomrule
\end{tabular}

\caption{Comparison between the standard PPO final checkpoint and the 1M-step variants considered in this work. PPO is evaluated at 2M environment steps, while the other methods are evaluated at 1M steps. The table illustrates that offline-guided training approaches the same operating regime as the final PPO policy while requiring approximately half of the online training budget. The SFT model is included for reference.}
\label{tab:final_checkpoints}

\end{table}

\begin{table}[t]
\centering
\small
\setlength{\tabcolsep}{5pt}
\renewcommand{\arraystretch}{1.1}
\begin{tabular}{lccccc}
\toprule
 & IND & \multicolumn{4}{c}{OOD} \\
\cmidrule(lr){3-6}
Method &  & Act & Lang & Vis & Avg \\
\midrule

PPO (1M)
& 0.75
& 0.67
& 0.62
& 0.65
& 0.64 \\

PPO SFT-init (1M)
& 0.87
& 0.51
& 0.69
& \textbf{0.78}
& 0.69 \\

\quad {\footnotesize $\Delta$ vs PPO}
& {\footnotesize +0.12}
& {\footnotesize -0.16}
& {\footnotesize +0.07}
& {\footnotesize +0.13}
& {\footnotesize +0.05} \\

PPO RefKL (1M)
& \textbf{0.93}
& \textbf{0.79}
& \textbf{0.76}
& 0.76
& \textbf{0.77} \\

\quad {\footnotesize $\Delta$ vs PPO}
& {\footnotesize +0.18}
& {\footnotesize +0.12}
& {\footnotesize +0.14}
& {\footnotesize +0.11}
& {\footnotesize +0.13} \\

PPO DataBC (1M)
& 0.91
& 0.74
& 0.73
& 0.74
& 0.74 \\

\quad {\footnotesize $\Delta$ vs PPO}
& {\footnotesize +0.16}
& {\footnotesize +0.07}
& {\footnotesize +0.11}
& {\footnotesize +0.09}
& {\footnotesize +0.10} \\

\bottomrule

\end{tabular}

\caption{Performance comparison at an equal reinforcement learning training budget of 1M steps. Both guided variants outperform PPO trained for the same number of steps, while PPO initialized from the SFT checkpoint improves over plain PPO but remains below the guided methods.}
\label{tab:main_results_1m}

\end{table}

We evaluate the proposed offline-guided variants within this controlled
experimental setting and compare them with standard PPO training in terms
of training efficiency and generalization across IND and OOD metrics.

Table~\ref{tab:main_results_1m} reports the results after 1M environment
steps. At this stage of training, both offline-guided variants outperform PPO across IND and all OOD categories.

The first obvious alternative is to initialize PPO directly from the
SFT LoRA and continue RL optimization from that point. Although this
strategy starts from substantially higher performance than standard PPO,
its subsequent improvement is surprisingly slow. As shown in
Table~\ref{tab:main_results_1m}, by 1M steps the SFT-initialized run
performs better than plain PPO at the same budget, but still remains
below both guided methods.
One possible explanation is that, after fitting the supervised
distribution more strongly, the LoRA adapter becomes harder to adapt
under RL updates alone, especially under distribution shift.

The training dynamics are shown in Fig.~\ref{fig:training_curves}.
Across both IND and OOD evaluations, the offline-guided variants reach
higher success earlier in training than standard PPO, though RefKL shows more consistently better performance than DataBC.

Table~\ref{tab:final_checkpoints} compares the resulting checkpoints.
While PPO reaches its final performance after 2M environment steps,
the offline-guided variants achieve comparable IND and OOD success
after 1M steps.

\subsection{Regularization Weight Ablations}

To study the effect of auxiliary supervision strength, we evaluate
multiple values of the initial regularization coefficient $\beta_0$.
These experiments follow the scheduling procedure described in
Section~\ref{sec:scheduling}, while varying only the initial magnitude of
the regularization term. The results are summarized in
Table~\ref{tab:beta_summary}.

In addition to the curriculum schedule, we also evaluate a
\textit{constant-$\beta$} variant where the coefficient remains fixed
throughout training. This comparison allows us to isolate the effect
of the curriculum from the overall strength of the auxiliary loss.
The table reports results for different values of $\beta_0$ under the
curriculum schedule and additionally includes the constant-$\beta$
configuration.

The results show that performance decreases as the initial coefficient
$\beta_0$ is reduced, for both IND and OOD evaluations, which is
expected in this setting. In other words, stronger auxiliary supervision leads to consistently higher
success rates across all settings.

As shown in Table~\ref{tab:beta_summary},
keeping $\beta$ constant does not improve final performance relative to the curriculum
schedule. Since the constant-$\beta$ variant also requires additional
forward passes through the reference model during the entire training
process, it introduces unnecessary computational overhead without
providing measurable benefits.
We also checked the interaction between PPO and auxiliary gradients and
did not observe noticeable gradient conflicts in practice.

\begin{table}[t]
\centering
\small
\setlength{\tabcolsep}{8pt}
\renewcommand{\arraystretch}{1.1}
\begin{tabular}{c|cc}
\hline
$\beta$ & IND & OOD (avg) \\
\hline
0.1 & $0.68$ & $0.52$ \\
0.3 & $0.82$ & $0.63$ \\
\textbf{0.6} & \textbf{0.93} & \textbf{0.77} \\
0.6 (no curriculum) & $0.86$ & $0.75$ \\
\hline
\end{tabular}
\caption{Effect of offline-guidance strength controlled by the coefficient $\beta$. Metrics are reported for RefKL PPO at 1M environment steps for both in-distribution and out-of-distribution evaluation. The final row additionally reports results for a constant $\beta$ schedule without curriculum decay.}
\label{tab:beta_summary}
\end{table}

\section{Analysis}

\subsection{Why Offline Guidance Improves Sample Efficiency}

The main empirical result of this work is that offline-guided RL reaches
the level of performance achieved by a much longer PPO run while using substantially fewer online
environment interactions. This effect is consistent across both IND and
OOD metrics. At 1M environment steps, both guided variants outperform
standard PPO at the same training budget, and RefKL already matches the
final 2M-step PPO checkpoint on average OOD success.

One plausible explanation is that the auxiliary offline objective
provides a strong behavioral prior during the early stage of RL training,
when the on-policy signal is still weak and exploration is inefficient.
Instead of learning useful action structure from sparse reward alone, the
policy starts from behavior that is already task-relevant and then adapts
it through online interaction. In this setting, offline supervision does
not replace RL, but accelerates the phase in which RL discovers how to
use the pretrained VLA representation effectively \cite{nair2020accelerating,lee2022offline2online,kang2018pofd}.

\subsection{Reference-Policy Guidance vs. Offline Behavior Cloning}

The clearest difference between the two guided variants appears in the
training dynamics. Across the learning curves, RefKL stays more
consistently above both standard PPO and DataBC, indicating that
reference-policy guidance provides a more reliable optimization signal
throughout training.

One possible reason is that the reference-policy objective is a smoother
regularizer. It preserves the action preferences encoded by supervised
training while still allowing PPO to adapt the policy through on-policy
optimization. In contrast, direct BC on a fixed offline dataset may
impose a stronger constraint toward the demonstration distribution,
making it harder to benefit from online improvement when the policy
encounters states that differ from the offline data.
In addition, for a large-scale pretrained model such as OpenVLA, the full
output distribution of the reference policy may carry richer information
than a single demonstrated action target.

\subsection{Discussion of SFT-Initialized RL}

An additional baseline considered in this work is to initialize PPO
directly from the SFT LoRA and continue RL training from that point.
This is the most direct way of combining offline and online learning,
and it indeed starts from a much stronger checkpoint than standard PPO.
However, the resulting training progress is much slower than in the
guided variants. By 1M steps, the SFT-initialized run remains below both
RefKL and DataBC despite outperforming plain PPO.

This slower adaptation is not uniform across OOD dimensions. In
particular, the SFT-initialized run is
relatively strong on vision and language OOD, but remains noticeably
weaker on action OOD. A plausible interpretation is that, once the LoRA
adapter has fit the supervised distribution more strongly, adapting its
action behavior under RL alone becomes harder, especially in settings
that require larger execution-level shifts. This makes direct
SFT-initialized RL a useful reference point, but not the main practical
solution in the current study.

\subsection{Limitations}

The conclusions of this work should be interpreted in the context of the
current experimental scope. First, several configurations are evaluated
with only one or two training seeds, so the results should be viewed as
strong empirical evidence rather than a definitive estimate of variance.
Second, the study is conducted within a single benchmark and model
family with LoRA-based OpenVLA adaptation. While this is
already a meaningful large-scale setting, further validation is needed
to determine how broadly the observed behavior transfers to other VLA
architectures, tasks, and reward formulations.

\section{Conclusion}

We study how offline supervision can be used as a structural prior during
RL fine-tuning of large-scale VLA policies. The results indicate that
offline guidance does more than provide a better initialization: when
incorporated into the RL objective, it stabilizes early optimization and
reduces the amount of online interaction required to reach strong policy
behavior. Among the variants considered here, reference-policy guidance is
the most effective and consistent formulation.

These results show that, in this setting, offline supervision and RL are
not competing alternatives but complementary signals. Supervised
knowledge from demonstrations helps stabilize and accelerate the early
phase of policy optimization, while RL remains essential for achieving
strong final performance, especially in OOD generalization. The best-performing configuration uses strong
initial guidance together with a curriculum that gradually removes the
auxiliary constraint, allowing the policy to transition to pure RL once
useful behavior has been established.

From a practical perspective, this improves the viability of RL fine-tuning:
the proposed hybrid training reduces the amount of expensive online
interaction required to reach strong performance, while preserving the
main advantages of RL over offline-only training. Future work should
evaluate this approach across a broader set of tasks and models, and
further analyze the mechanisms by which offline supervision improves
online optimization.

\bibliography{main}
\bibliographystyle{icml2026}




\end{document}